\documentclass[a4paper]{article}
\usepackage[letterpaper,margin=0.75in]{geometry}
\usepackage{spconf,amsmath,graphicx}
\usepackage{amssymb}
\usepackage{amsfonts}
\usepackage{color}
\usepackage{multirow}
\usepackage[all,poly]{xy}
\usepackage{algorithm}
\usepackage{algorithmic}
\usepackage{subfig}
\usepackage{cite}
\usepackage{environ}

\title{Patch-based Sparse Representation For Bacterial Detection}
\name{A. K. Eldaly$^{(1), (2)}$, Y. Altmann$^{(1)\dagger}$\thanks{This work was supported in parts by the $^\dagger$Royal Academy of Engineering through the research fellowship scheme (RF201617/16/31) and $^*$the EPSRC via grant EP/K03197X/1.}, A. Akram$^{(2)}$, A. Perperidis$^{(1), (2)*}$, K. Dhaliwal$^{(2)}$, S. McLaughlin$^{(1)}$}
\address{$^{(1)}$School of Engineering and Physical Sciences, Heriot-Watt University, Edinburgh, United Kingdom\\ 
$^{(2)}$The Queen's Medical Research Institute, The University of Edinburgh, Edinburgh, United Kingdom} %\\
\def\bs0{{\boldsymbol{0}}}

\def\bfPsi{{\boldsymbol{\Psi}}}
\def\bfUpsilon{{\boldsymbol{\Upsilon}}}

\def\bfy{{\mathbf{y}}}

\def\bfD{{\mathbf{D}}}

\def\bfF{{\mathbf{F}}}

\def\bfI{{\mathbf{I}}}

\def\bfM{{\mathbf{M}}}

\def\bfR{{\mathbf{R}}}

\def\bfW{{\mathbf{W}}}

\def\bfY{{\mathbf{Y}}}
\def\bfZ{{\mathbf{Z}}}

\newcommand{\norm}[1]{\left\|#1\right\|}

\newcounter{algo}
\renewcommand{\thealgo}{\arabic{algo}}

\usepackage[font=small,skip=2pt]{caption}

\begin{document}

\ninept
\maketitle
\setlength{\abovedisplayskip}{1pt}
\setlength{\belowdisplayskip}{1pt}
\setlength{\belowcaptionskip}{-13pt}

\begin{abstract}
\vspace*{-0.0cm}
In this paper, we propose an unsupervised approach for bacterial detection in optical endomicroscopy images. This approach splits each image into a set of overlapping patches and assumes that observed intensities are linear combinations of the actual intensity values associated with background image structures, corrupted by additive Gaussian noise and potentially by a sparse outlier term modelling anomalies (which are considered to be candidate bacteria). The actual intensity term representing background structures is modelled as a linear combination of a few atoms drawn from a dictionary which is learned from bacteria-free data and then fixed while analyzing new images. The bacteria detection task is formulated as a minimization problem and an alternating direction method of multipliers (ADMM) is then used to estimate the unknown parameters. Simulations conducted using two \emph{ex vivo} lung datasets show good detection and correlation performance between bacteria counts identified by a trained clinician and those of the proposed method.

\end{abstract}

%\graphicspath{{figures/}}
\vspace*{-0.1cm}
\section{Introduction}
\label{sec:Introduction}
\vspace*{-0.1cm}
Outlier/anomaly detection problems can usually be addressed using unsupervised or supervised methods \cite{smal2010quantitative}. In unsupervised approaches, the objects/anomalies to be detected are learned from the data by fitting them with suitable distributions without using explicitly-provided labels \cite{ding2011bayesian, Altmann2015a, mccoolMCMC2016, smal2008multiple, smal2008new, bright1987two, kimori2010extended, he2010laplacian, Eldaly2019Bayesian}. On the other hand, considering supervised approaches, the dataset is usually divided into training and testing sets. In the training phase, a model is trained by pairing inputs with their expected outputs, which are also known as the ground truth. The trained model can then be used to estimate the output of the test dataset \cite{seth2018estimating, jiang2007detection, arteta2016counting, lempitsky2010learning}.

In this work, we investigate the performance of an unsupervised approach for bacterial detection in datasets of optical endomicroscopy (OEM) lung images \cite{perperidis2018image, ayache2006processing, le2004towards, krstajic2016two, eldaly2017deconvolution, eldaly2018deconvolution, perperidis2017characterization}. The main contributions of this work are threefold. First, we formulate the problem of simultaneous bacteria detection and background estimation as a (robust) sparse coding problem and use an ADMM algorithm to solve the bacteria detection problem. To the best of our knowledge, it is the first time this problem is addressed by a sparse representation approach. Second, we provide simulations using real datasets, whereby we investigate different bacterial concentrations including control cases in which no bacteria are present, and different SmartProbes that cause weak and strong bacteria fluorescence. Third, we compare the results of the proposed model with bacteria annotations performed by a trained clinician and three widely used spot-detection algorithms, using both dot-annotation and count-annotation methods.

The reminder of the paper is organized as follows. Section \ref{sec:ProbelmFormulationDetectionPatchBased} formulates briefly the problem of bacteria detection in OEM images using a sparse coding approach, followed by Section \ref{sec:ProposedModel}, which summarizes the proposed model and possible estimation strategy to recover the unknown model parameters. Experimental results and discussions are presented in Section \ref{sec:RealDataResultsDetectionPatchBased}. Conclusions and future work are finally reported in Section \ref{sec:Conclusion}.
\vspace*{-0.1cm}
\section{Outlier Detection Formulation}
\label{sec:ProbelmFormulationDetectionPatchBased}
\begin{figure}
\centering  
\subfloat{\label{subfig:frame1_13}\includegraphics[scale = 0.27]{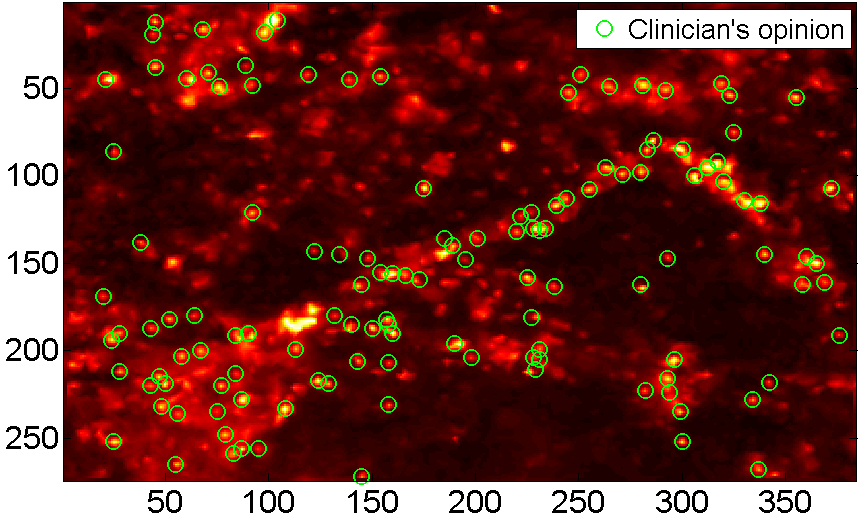}}
\caption{An OEM image with bacteria shown within circles annotated by a trained clinician.}
\label{fig:ProbForm}
\end{figure}

\vspace*{-0.1cm}
Figure \ref{fig:ProbForm} shows an example of an OEM image with bacteria shown within circles that are annotated by a trained clinician. We can observe that bacteria appear as high intensity dots in the image in addition to as bright as background structures representing elastin and collagen, making the differentiation of bacteria a quite challenging task. The problem of bacteria detection is formulated such that, given a test image $\mathcal{I} \in \mathbb{R}^{m \times n}$, a data matrix $\bfY \in \mathbb{R}^{P\times L}$ is formed by splitting the image into a set of $L$ overlapping square patches containing $P=N_p^2$ pixels. These patches are vectorized and finally gathered in $\bfY = [\bfy_1, \bfy_2, \cdots, \bfy_L]$. The data matrix $\bfY$ can then be well approximated by a sparse linear model, excluding a small number of pixels - the outliers - which significantly deviate from this model. The collection $\bfY$ is described as follows

\begin{equation}
\begin{aligned}
\bfY = \bfD \bfPsi + \bfR + \bfW,
\label{eq:dicLearning4}
\end{aligned}
\end{equation}
where $\bfD \in \mathbb{R}^{P \times K}$ is a dictionary assumed to be known, $\bfPsi \in \mathbb{R}^{K \times L}$ is the sparse coefficient matrix, $\bfR \in \mathbb{R}^{P \times L}$ has few non-zero elements that represents sparse deviations from the linear model $\bfD\bfPsi$ representing background image structures, and $\bfW \in \mathbb{R}^{P \times L}$ is a low-energy noise component, which is assumed to be independent and identically distributed (i.i.d.) Gaussian.

The primary objective here is to estimate the outlier matrix $\bfR$ in Eq. \eqref{eq:dicLearning4}, given that the sparse coefficients in $\bfPsi$ are also unknown. Thus we proposes to estimate jointly ($\bfR$, $\bfPsi$) from the observation matrix $\bfY$. To solve this problem, we propose an optimization-based method to estimate the unknown parameters.
\vspace*{-0.2cm}
\section{Proposed Model}
\label{sec:ProposedModel}
\vspace*{-0.2cm}
The recovery of $\bfR$ and $\bfPsi$ in Eq.\eqref{eq:dicLearning4} is formulated as the following unconstrained minimization problem
\begin{equation}
\begin{aligned}
%\{\hat{\bfPsi}, \hat{\bfR}\} = 
\underset{\bfPsi, \bfR}{\text{minimize}}
& & \frac{1}{2}\norm{\bfY - \bfD\bfPsi - \bfR}^2_F + \alpha \norm{\bfPsi}_{1, 1} + \beta \norm{\bfR}_{1, 1},
\label{eq:ProblemFormulation0}
\end{aligned}
\end{equation}
where $\bfPsi = [\bfPsi_1, \cdots \bfPsi_K]^T$, $\norm{\bfPsi}_{1, 1} = \sum_{k}\norm{\bfPsi(k, :)}_{1} $, similarly $\bfR = [\bfR_1, \cdots \bfR_P]^T$, $\norm{\bfR}_{1, 1} = \sum_{p}\norm{\bfR(p, :)}_{1} $, $\alpha$ and $\beta$ are two positive scalar parameters controlling the degree of sparsity of $\bfPsi$ and $\bfR$ respectively, and $\norm{\cdot}_F$ denotes the Frobenius norm. Problem \eqref{eq:ProblemFormulation0} encourages a solution in which $\bfPsi$ is sparse. However, for the outliers that cannot be represented exclusively by $\bfD$, it permits non-zero entries in $\bfR$.

The optimization problem in Eq. \eqref{eq:ProblemFormulation0}, although convex, cannot be solved using standard gradient-based methods due to the non-smooth terms. The core idea is to convert this unconstrained minimization problem into another constrained one by the application of a variable splitting operation (see Eq. \eqref{eq:ProblemFormulation1} below). Finally, the obtained constrained problem is solved with using ADMM \cite{afonso2011augmented, nocedal2006numerical, eldaly2017deconvolution}. By a careful choice of the new variables, the initial problem is converted into a sequence of much simpler problems, which can be solved iteratively. To solve the problem depicted in Eq. (\ref{eq:ProblemFormulation1}), we introduce a new variable $\bfZ$ for the regularization function in $\bfPsi$ in order to decouple it from the data fidelity term. Therefore, the constrained version of problem \eqref{eq:ProblemFormulation0} can be written as follows
\begin{eqnarray}
%\{\hat{\bfPsi}, \hat{\bfR}\} = 
&\underset{\bfPsi, \bfR}{\text{minimize  }} \frac{1}{2}\norm{\bfY - \bfD\bfPsi - \bfR}^2_F + \alpha \norm{\bfZ}_{1, 1} + \beta \norm{\bfR}_{1, 1},&\nonumber\\
&\text{subject to} \hspace{0.35cm} \bfZ = \bfPsi.&
\label{eq:ProblemFormulation1}
\end{eqnarray} 

The augmented Lagrangian corresponding to the problem in Eq. \eqref{eq:ProblemFormulation1} can be written as $\mathcal{L}(\bfPsi, \bfR, \bfZ, \bfM) = \frac{1}{2}\norm{\bfY - \bfD\bfPsi - \bfR}^2_F + \alpha \norm{\bfZ}_{1, 1}
+ \beta \norm{\bfR}_{1, 1} + \frac{\mu}{2}\norm{\bfZ - \bfPsi - \bfM}^2_F$, 
%\begin{eqnarray}
%\mathcal{L}(\bfPsi, \bfR, \bfZ, \bfM) & = & \frac{1}{2}\norm{\bfY - \bfD\bfPsi - \bfR}^2_F + \alpha \norm{\bfZ}_{1, 1}\nonumber\\
%& + &  \beta \norm{\bfR}_{1, 1} + \frac{\mu}{2}\norm{\bfZ - \bfPsi - \bfM}^2_F,
%\label{eq:ProblemFormulation3}
%\end{eqnarray} 
where $\bfM$ is the set of Lagrange multiplier corresponding to the splitting, and $\mu > 0$ is a constant. The ADMM algorithm using to solve Eq. \eqref{eq:ProblemFormulation1} (also Eq. \eqref{eq:ProblemFormulation0}) is shown in Algorithm \ref{algo:ADMMSparseCoding}. During each step of this iterative scheme, $\mathcal{L}$ is optimized with respect to $\bfPsi$ (step 2), $\bfR$ (step 3) and $\bfZ$ (step 4), and then the Lagrange multipliers are updated (step 5). %The algorithm terminates when a stopping criterion is satisfied, as will be mentioned below.
\begin{algorithm}
\caption{Sparse coding with Bacterial Detection - Version I}
\label{algo:ADMMSparseCoding}
\begin{algorithmic}[1]
\item set $k = 0$, choose $\mu > 0, \bfPsi^{(0)}, \bfR^{(0)}, \bfZ^{(0)}, \text{ and } \bfM^{(0)}$
\STATE \textbf{repeat} ($k \leftarrow k + 1$)
\STATE $\bfPsi^{(k+1)} \leftarrow \text{minimize}_\bfPsi \, \mathcal{L}(\bfPsi, \bfR^k, \bfZ^k, \bfM^k)$
\STATE $\bfR^{(k+1)} \leftarrow \text{minimize}_\bfR \, \mathcal{L}(\bfPsi^{k+1}, \bfR, \bfZ^k, \bfM^k)$
\STATE $\bfZ^{(k+1)} \leftarrow \text{minimize}_\bfZ \, \mathcal{L}(\bfPsi^{k+1}, \bfR^{k+1}, \bfZ, \bfM^{k})$
\STATE $\textbf{Update } \bfM: $ $\bfM^{(k+1)} \leftarrow \bfM^{(k)} - \left(\bfZ^{(k+1)} - \bfPsi^{(k+1)}\right)$
\STATE \textbf{until} some stopping criterion is satisfied.
\end{algorithmic}
\end{algorithm}
\vspace*{-0.3cm}

Solving the minimization problems in Algorithm \ref{algo:ADMMSparseCoding} leads to Algorithm \ref{algo:ADMMSparseCoding1}, where $\bfUpsilon = \bfY - \bfR$, $\bfF = \bfZ + \bfM$ and \textit{soft} is the soft thresholding function \cite{combettes2005signal}. The parameter $\mu > 0$ is updated within the algorithm to keep the primal and dual residual norms within a factor of $10$ of one another. The stopping criterion we use is $\left(\norm{\bfZ^{(k)} - \bfPsi^{(k)}}_F + \mu\norm{\bfM^{(k)} - \bfM^{(k+\rho)}}_F\right) \leq \epsilon $, which is the sum of the primal and dual residuals, where $\epsilon = \sqrt{P \times L}\times 10^{-6}$ \cite{afonso2011augmented, eldaly2017deconvolution}.
\begin{algorithm}
\caption{Sparse coding with Bacterial Detection - Version II}
\label{algo:ADMMSparseCoding1}
\begin{algorithmic}[1]
\STATE set $k = 0$, choose $\mu > 0, \bfPsi^{(0)}, \bfR^{(0)}, \bfZ^{(0)}, \text{ and } \bfM^{(0)}$
\STATE \textbf{repeat} ($k \leftarrow k + 1$)
\STATE $\bfPsi^{(k+1)} \leftarrow (\bfD^T\bfD + \mu^{(k)} \bfI)^{-1} (\bfD^T\bfUpsilon^{(k)} + \mu^{(k)} \bfF^{(k)})$,
\STATE $\bfR^{(k+1)} \leftarrow \text{soft} (\bfY - \bfD\bfPsi^{(k+1)}, \beta)$,
\STATE $\bfZ^{(k+1)} \leftarrow \text{soft} (\bfPsi^{(k+1)} - \bfM^{(k)}, \frac{\alpha}{\mu})$,
\STATE $\bfM^{(k+1)} \leftarrow \bfM^{(k)} - (\bfZ^{(k+1)} - \bfPsi^{(k+1)})$ 
\STATE \textbf{until} some stopping criterion is satisfied.
\end{algorithmic}
\end{algorithm}
\vspace*{-0.4cm}
\section{Experimental Results}
\label{sec:RealDataResultsDetectionPatchBased}
\vspace*{-0.2cm}
\subsection{Datasets}
\vspace*{-0.1cm}
The proposed algorithm was assessed using two datasets of \emph{ex vivo} ventilated whole ovine lungs with bacteria present. Dataset I contains seven videos assessing a combination of fluorescent dyes (SmartProbes) and bacterial types, including control segments. It contains (i) three videos of ovine lungs instilled with \emph{Methicillin-sensitive Staphylococcus aureus} (\emph{MSSA}) stained with a commercially available laboratory dye (PKH67, Sigma-Aldrich), a highly fluorescent cell membrane dye, (ii) two videos of ovine lungs instilled with bacteria (gram-positive \emph{MSSA} and gram-negative \emph{Pseudomonas} PA3284) stained in situ with an in-house bacterial detection SmartProbe \cite{akram2016structural}, and (iii) two videos of ovine lungs without the presence of any bacteria. Videos 1 to 5 were instilled with a single concentration of bacteria, equivalent to Optical Density (OD595nm) of 2. 

\begin{table}
\centering
\begin{tabular}{|c|c|c|c|c|}
\hline
\textbf{Video} & \textbf{\begin{tabular}[c]{@{}c@{}}\# of \\ frames\end{tabular}} & \textbf{\begin{tabular}[c]{@{}c@{}}Bacteria \\ concent.\\ (OD)\end{tabular}} & \textbf{Fluorophore} & \textbf{Bacteria} \\ \hline
\textbf{1} & 26 & \multirow{5}{*}{2} & \multirow{3}{*}{PKH} & \multirow{3}{*}{\textit{\begin{tabular}[c]{@{}c@{}}Staphylococcus\\ aureus\end{tabular}}} \\ \cline{1-2}
\textbf{2} & 19 &  &  &  \\ \cline{1-2}
\textbf{3} & 13 &  &  &  \\ \cline{1-2} \cline{4-5} 
\textbf{4} & 32 &  & \multirow{3}{*}{SmartProbe} & \textit{\begin{tabular}[c]{@{}c@{}}Pseudomonas\\ aeruginosa\end{tabular}} \\ \cline{1-2} \cline{5-5} 
\textbf{5} & 19 &  &  & \textit{\begin{tabular}[c]{@{}c@{}}Staphylococcus\\ aureus\end{tabular}} \\ \cline{1-4} \cline{5-5} 
\textbf{6} & 12 & \multirow{2}{*}{NA} & \multirow{2}{*}{NA} & \multirow{2}{*}{Control} \\ \cline{1-2}
\textbf{7} & 12 &  &  &  \\ \hline
\end{tabular}
\caption{Description of dataset I.}
\label{tab:data_set}
\end{table}

Dataset II contains four videos, each with an increasing bacterial concentration (OD595nm 0.004, 0.04, 0.4, 4), all labelled with an in-house bacterial detection SmartProbe. This dataset is considered to make sure that as the concentration increases, the counts of the clinician and of the algorithm also increase. Tables \ref{tab:data_set} and \ref{tab:data_set_2} summarise the details of Datasets I and II respectively. 

\begin{table}[!h]
\centering
\begin{tabular}{|c|c|c|c|c|}
\hline
\textbf{Video} & \textbf{\begin{tabular}[c]{@{}c@{}}\# of \\ frames\end{tabular}} & \textbf{\begin{tabular}[c]{@{}c@{}}Bacteria\\ concent.\\ (OD)\end{tabular}} & \textbf{Fluorophore} & \textbf{Bacteria} \\ \hline
\textbf{1} & 14 & 0.004 & \multirow{4}{*}{SmartProbe} & \multirow{4}{*}{\textit{\begin{tabular}[c]{@{}c@{}}Pseudomonas\\ aureus\end{tabular}}} \\ \cline{1-3}
\textbf{2} & 14 & 0.04 &  &  \\ \cline{1-3}
\textbf{3} & 15 & 0.4 &  &  \\ \cline{1-3}
\textbf{4} & 15 & 4 &  &  \\ \hline
\end{tabular}
\caption{Description of dataset II.}
\label{tab:data_set_2}
\end{table}

The Cellvizio fibred confocal OEM imaging platform (Mauna Kea Technologies, Paris, France) \cite{ayache2006processing, le2004towards} was used to acquire all data in this study. Image sequences of size $274 \times 384$ pixels ($306\mu m \times 429\mu m$) were captured at 12 frames per second. Random frames that are representative of each of the entire video sequences are chosen from each of the two datasets by a trained clinician. These comprise 133 image frames for Dataset I, and 58 frames for Dataset II as described in Tables \ref{tab:data_set} and \ref{tab:data_set_2} respectively. In each frame, a trained clinician marked the coordinates of phenomena that are thought with high confidence to be bacteria. Ambiguous points are ignored. 
\vspace*{-0.2cm}
\subsection{Dictionary Learning for Bacterial Detection}
\begin{figure}
\centering
\includegraphics[width=0.9\linewidth]{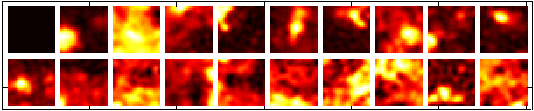}
\caption{Example of dictionary atoms learned from Dataset I.}
\label{fig:learntDictionary1}
\end{figure}

\vspace*{-0.1cm}
Each dataset is split into training and testing phases. In the training phase, one dictionary is learned for each dataset from its corresponding videos; namely $\textbf{D}_1$ for Dataset I and $\textbf{D}_2$ for Dataset II. Every set of frames in each video has a certain elastin and collagen pattern. Hence, one frame from each set is chosen as a representative. This yielded 12 frames from Dataset I and 17 frames from Dataset II. Features are then extracted from each frame by dividing it into square overlapping patches of fixed size. In this work, we employed a $27 \times 27$ window size with $50 \%$ overlap. The patches that are annotated by the clinician as containing bacteria are then excluded from the training dataset (see Fig. \ref{fig:ProbForm}). The remaining bacteria-free patches are vectorised and gathered in the training matrix $\bfY_{tr} \in \mathbb{R}^{729 \times 3553}$ for Dataset I and $\bfY_{tr} \in \mathbb{R}^{729 \times 7087}$ Dataset II. The method of optimal directions (MOD) dictionary learning method \cite{engan2000multi} is then applied to train the dictionaries. The K-SVD algorithm \cite{aharon2006k} is also investigated but provided similar results to MOD, thus the results are not reported here. Figure \ref{fig:learntDictionary1} shows 20 dictionary atoms learned for a selection of frames from Datasets I.

\vspace*{-0.3cm}
\subsection{Algorithm evaluation}
\label{subsubsec:PostProc1}
\vspace*{-0.1cm}
In the testing phase, after the dictionaries have been learned, Algorithm \ref{algo:ADMMSparseCoding1} is run for each of the remaining frames which are $111$ (resp. $41$) remaining frames for Dataset I (resp. Dataset II), yielding the estimated outlier matrix $\hat{\bfR}$ for each frame. The final outlier image is then reconstructed using these overlapping patches by averaging their intensities. The outlier image is then normalized to $[0,1]$ range and thresholded (by $\ell_d$), while pixels that exceed this threshold value are counted as a potential bacteria. Since each bacterium corresponds to a set of connected pixels, each group of connected detections is counted as a single detection. The estimated number of bacteria is thus the number of estimated groups and their positions are computed using the barycenter of each region. 

Once these connected detections are identified, they are then replaced by a single detection at the mean of their locations, which gives the final number of detected bacteria in each frame. 

Due to the unbalance of this two-class problem (absence/ existence of bacteria), we consider precision-recall curves to assess the bacteria detection performance, in which the reference is the set of annotations from the clinicians. Precision-recall curves are plots of precision versus recall at different cut-off thresholds (different $\ell_d$) for the resulting outlier amplitude image. The precision and the recall can be calculated as $ \text{Precision} = \frac{\text{TP}}{\text{TP} + \text{FP}}, \hspace{ 0.1cm} \hspace{ 0.3cm} \text{Recall} = \frac{\text{FP}}{\text{FP} + \text{FN}}$ respectively,
where TP, FN, and FP refer to the number of true positives, false negatives, and false positives respectively. Given the pixel locations where a bacterium has been annotated by the clinician, we defined a disk of radius $r = 10$ pixels \cite{mandula2014localisation}, and we consider that any detection that is present within the disk as a match (TP); any detection outside any of the disks as FP; and any clinician's annotation that does not match with any of the algorithm detection as FN.

We test different parameters for evaluating the performance of the proposed algorithm. First, we fix the regularization parameter $\alpha$ corresponding to the sparse representation matrix $\bfPsi$ to $\alpha = 1\times 10^{-5}$, and vary the outlier regularization parameter ($\beta$). Second, we investigate the impact of the number of atoms ($K$) within the learned dictionary. Finally we vary the outlier amplitude image threshold ($\ell_d$) between $0$ and $1$, and construct the precision-recall curves accordingly. Statistical comparison of bacterial counts (count-annotation) and detections (dot-annotation) performed by the trained clinicians and the algorithm output is then considered after choosing the best combination of the parameters described above.
\vspace*{-0.3cm}
\subsection{Results and Discussion}
\label{sec:DiscussionPatchBased}
\begin{figure}
\centering  
\subfloat []{\label{subfig:Dataset1_Max_AUC_New3}\includegraphics[scale = 0.19]{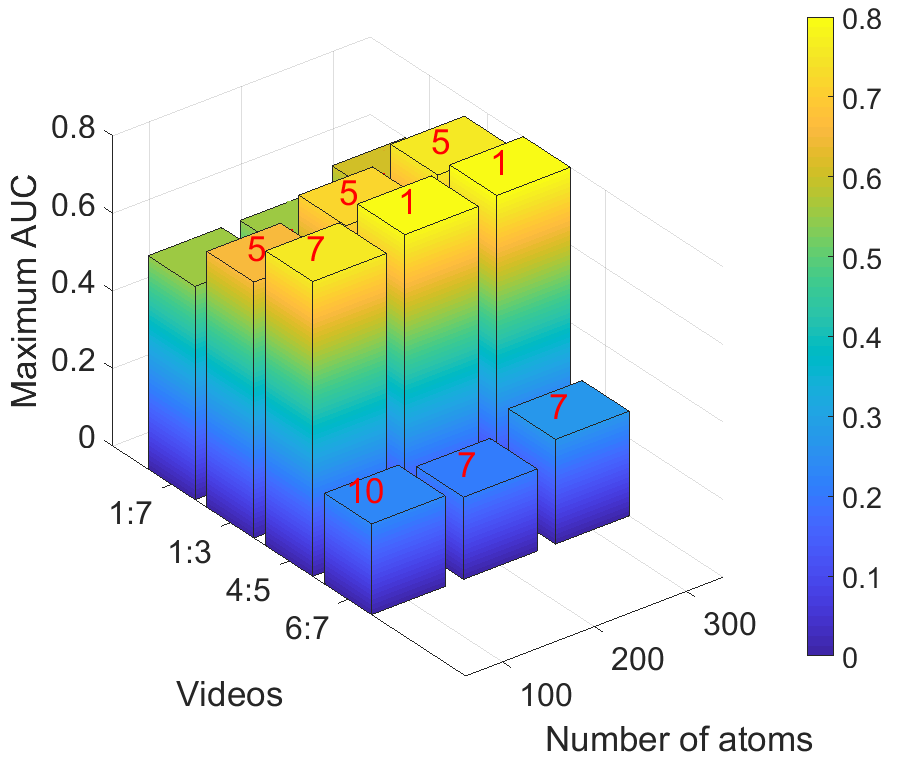}}
\subfloat []{\label{subfig:Dataset2_Max_AUC_New3}\includegraphics[scale = 0.19]{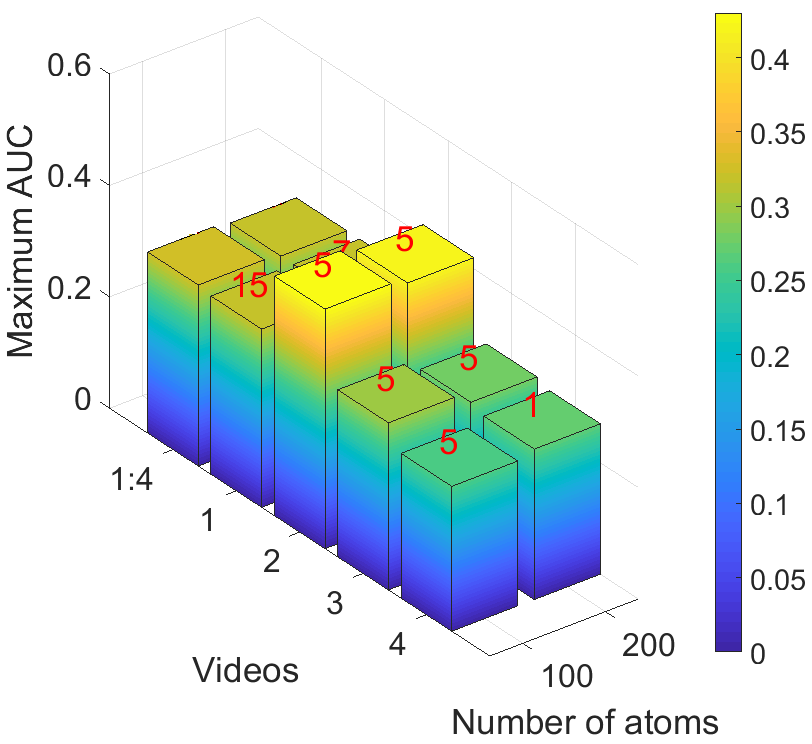}}
\caption{Plot of maximum achieved AUC reported for Dataset I in (a) and II in (b). The corresponding values of $\beta$ are provided above each bar in red.}
\label{fig:Dataset_Max_AUC_New}
\end{figure}

\vspace*{-0.1cm}
\textbf{Dot-annotation effect:} Figure \ref{fig:Dataset_Max_AUC_New}(a) shows a plot of different smartprobes (represented by video ranges) versus different numbers of dictionary atoms and the maximum achieved area under precision-recall curve (AUC). It can be noted that the bacteria detection performance is enhanced when increasing the number of dictionary atoms. Although a strong smartprobe which produces high fluorescence signals is used for videos 1:3, the reported AUC is close to that for videos 3:4, for which a weaker smartprobe is used. This is because videos 4 and 5 have less elastin and collagen structures, and hence there is lower probability of getting false positive detections. Regarding the control cases (videos 6 and 7), it can be observed that the optimal regularization parameter $\beta$ (printed in red on top of each bar) is always higher than that of the bacteria stained videos (videos 1:3 and 4:5), which in turn promotes more outlier sparseness and hence fewer counts. Moreover, the AUCs of these videos are lower than those of videos 1:3 and 4:5, as they are not stained by fluorophores and hence makes the fluorescence of bacteria weaker and more difficult to discriminate, stressing the need for SmartProbes for bacterial detection. We also noticed that there a broad range of outlier regularization parameters provides very similar precision-recall curves, and the results are not extremely sensitive to the value of $\beta$. 

Figure \ref{fig:Dataset_Max_AUC_New}(b), on the other hand, shows a plot of the four concentrations (represented by video numbers) versus different numbers of atoms and the maximum achieved AUC. We note that there is not much difference in the achieved maximum AUC for the two tested dictionary atom numbers. We also noticed that there is a broad range of the regularization parameter $\beta$ values that provides same AUC. % We can also observe that there is a general trend that when the bacteria concentration increases, the outlier regularization parameter $\beta$ value decreases.

\begin{figure}
\centering
\includegraphics[width=0.8\linewidth]{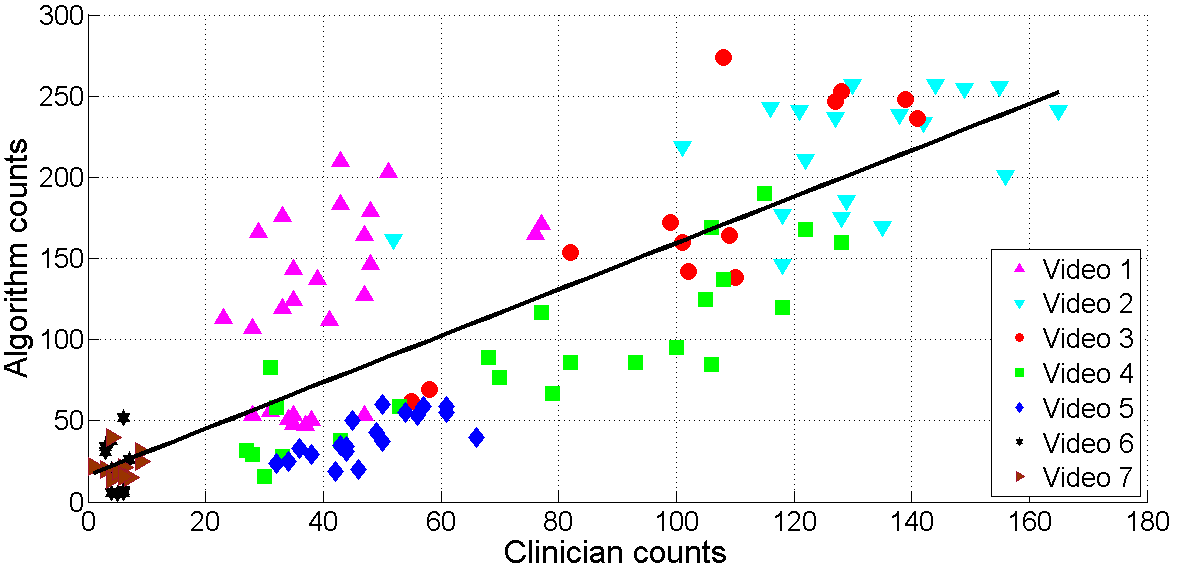}
\caption{Plot of clinician bacteria count versus algorithm bacteria count for Dataset I. Dots correspond to frames, and colours correspond to videos.}
\label{fig:Counts}
\end{figure}

\textbf{Count-annotation effect:} For Dataset I, the algorithm counts are compared with the clinician counts in each frame as shown in Fig. \ref{fig:Counts}. This corresponds to precision of $50\%$ and recall of $86.12\%$ that also corresponds to cut-off threshold to the outlier amplitude images of $\ell_d = 0.07$. We considered the values of $\beta$ providing the maximum AUC per each fluorophore. We can observe an almost linear relationship between the clinician counts and algorithm counts, with an empirical linear correlation between the manually and automatically detected anomalies as $0.823$. Furthermore, for videos 1, 2 and 3 in which a highly fluorescent SmartProbe is used, and videos 4 and 5 in which an in-house SmartProbe which produces weaker fluorescence signals is used, a similar trend is observed between the numbers of clinician's annotations and the counts provided by our algorithm. This also depending on the type of bacteria the samples are stained with. Videos 6 and 7, which are controls, show minimal annotations and counts, which reflects the ability of the algorithm to differentiate bacterial loads from control. % Moreover, the algorithm provides low precision values as each frame

\begin{figure}
\centering  
\subfloat []{\label{subfig:AhsancountADMM}\includegraphics[scale = 0.19]{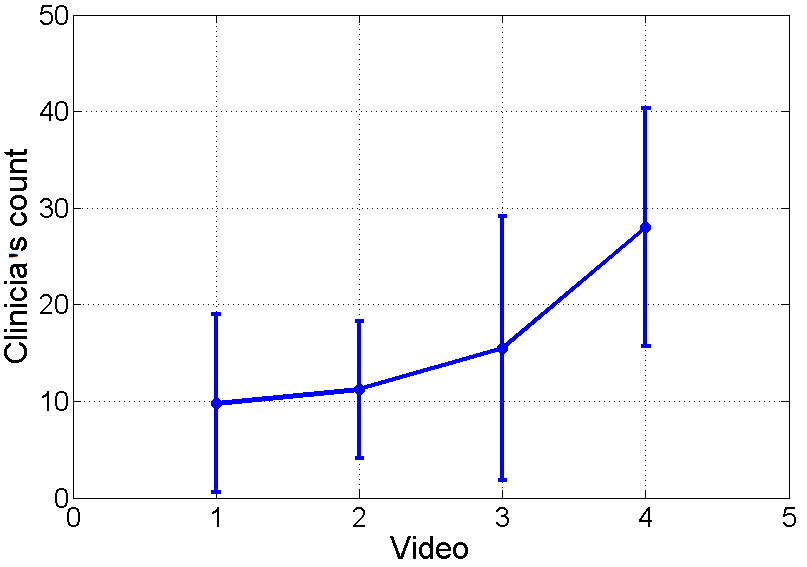}}
\subfloat []{\label{subfig:ADMMcount}\includegraphics[scale = 0.19]{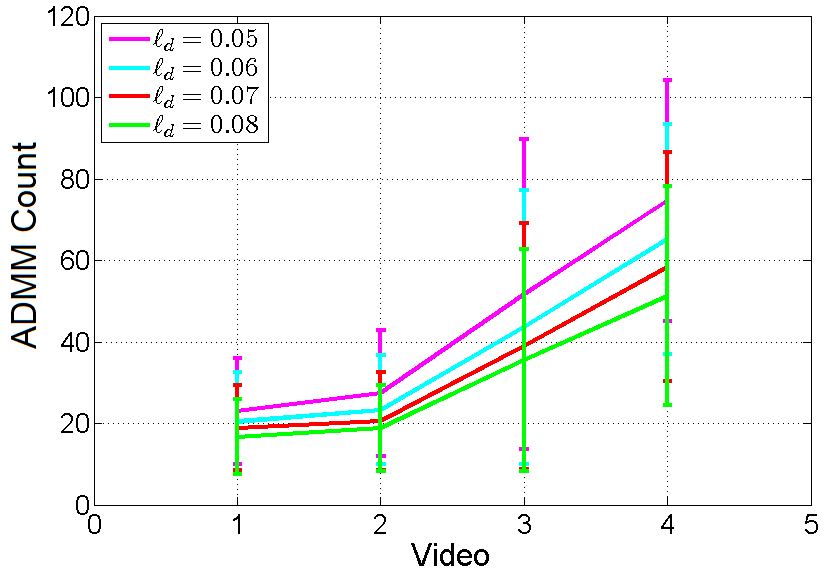}}
\caption{Mean number of detections per selected frames in videos 1 to 4 of Dataset II and the corresponding standard deviation. (a) clinician's opinion, (b) proposed method.}
\label{fig:IncreasingConcentADMM}
\end{figure}

Similarly, for Dataset II, we compared the clinician-algorithm counts for different cut-off thresholds $\ell_d$, ranging between $\ell_d = 0.05$ and $\ell_d = 0.08$, which corresponds to total sensitivity of $82.03\% \text{ to } 66.58\%$ and precision of $23.64\% \text{ to } 32.4\%$, and provided the results in Fig. \ref{fig:IncreasingConcentADMM}. This corresponds to the counts provided by maximum AUC when different values of $\beta$ are tested. We can observe that the counts of both the algorithm and the clinician increase as the bacteria concentration increases, which reflects the agreement between the approach considered and the clinician's annotations. 

We can also observe that the algorithm counts are higher than those of the clinician for the two processed datasets, as we expect the algorithm to be able to identify dots that are barely visible to the naked eye. Moreover, the clinician did not annotate ambiguous dots, meaning that a number of these were not chosen. This, along with false positives, is the main reason why the algorithm counts are higher than the clinician counts.
\vspace*{-0.3cm}
\subsection{Comparison with existing approaches}
\vspace*{-0.1cm}
In this subsection, we compare the proposed approach with popular spot-detection methods from the literature, namely the Laplacian of Gaussian (LoG) and its approximation; the difference of Gaussians (DoG) filters \cite{lindeberg1998feature, he2010laplacian}, and the grey scale opening top-hat filter (GSOTH) \cite{kimori2010extended, bright1987two}. These methods, although simple, have been considered in the literature of spot and blob detection in various applications. In this work, from preliminary trials to optimize performance, the LoG filter is implemented by employing a $5 \times 5$ kernel of standard deviation of $0.8$ to each frame. Similarly, the DoG filter is implemented by considering the difference of two $5 \times 5$ Gaussian kernels of standard deviations of $0.5$ and $0.8$ respectively. The GSOTH is employed by first smoothing the input image by a Gaussian kernel to reduce the noise, then by computing the morphological opening of the input image by employing a $3 \times 3$ flat disc, which achieves the best detection results and then subtracts the result from the original image. The same post processing steps described earlier (pixel grouping and computation of the barycenters) are also employed. The comparison is conducted in terms of AUC of the resulting precision-recall curves, as well as in terms of computation time.

Table \ref{tab:Comparison1} compares the maximum achieved AUC of the proposed algorithm for Datasets I and II with those of the three methods described above. We can observe that the proposed algorithm provides the highest AUC for both Datasets I and II. Although the grey scale opening top-hat filter provides competitive results for videos 1:3 and 4:5 in Dataset I, it fails to identify the control cases as good as the proposed approach. The LoG and the DoG filters, on the other hand, show similar performance.

\begin{table}[]
\centering
\begin{tabular}{cc|c|c|c|c|}
\cline{3-6}
 &  & \textbf{\begin{tabular}[c]{@{}c@{}}Sparse\\ coding\end{tabular}} & \textbf{LoG} & \textbf{DoG} & \textbf{\begin{tabular}[c]{@{}c@{}}GSOTH\end{tabular}} \\ \cline{2-6} 
\multicolumn{1}{c|}{} & Videos & \multicolumn{4}{c|}{AUC} \\ \hline
\multicolumn{1}{|c|}{\multirow{4}{*}{\textbf{Dataset I}}} & 1:3 & \textbf{0.754} & 0.58 & 0.56 &\underline{0.749} \\ \cline{2-6} 
\multicolumn{1}{|c|}{} & 4:5 & \textbf{0.8} & 0.53 & 0.63 & \underline{0.786} \\ \cline{2-6} 
\multicolumn{1}{|c|}{} & 6:7 & \textbf{0.27} & \underline{0.175} & 0.104 & 0.172 \\ \cline{2-6} 
\multicolumn{1}{|c|}{} & \textbf{Average} & \textbf{0.61} & 0.43 & 0.43 & \underline{0.569} \\ \hline
\multicolumn{1}{|c|}{\multirow{5}{*}{\textbf{Dataset II}}} & 1 & \textbf{0.32} & 0.142 & 0.14 & \underline{0.257} \\ \cline{2-6} 
\multicolumn{1}{|c|}{} & 2 & \textbf{0.43} & 0.18 & 0.268 & \underline{0.322} \\ \cline{2-6} 
\multicolumn{1}{|c|}{} & 3 & \textbf{0.30} & 0.09 & 0.116 & \underline{0.184} \\ \cline{2-6} 
\multicolumn{1}{|c|}{} & 4 & \textbf{0.26} & 0.136 & 0.115 & \underline{0.226} \\ \cline{2-6} 
\multicolumn{1}{|c|}{} & \textbf{Average} & \textbf{0.33} & 0.137 & 0.16 & \underline{0.247} \\ \hline
\end{tabular}
\caption{Area under curve measures of the resulting precision-recall curves of the proposed approach and three existing methods. Bold (resp. underlined) represent best (resp. second best) results.}
\label{tab:Comparison1}
\end{table}

The average computation times of the four methods are 0.4, 0.11, 0.05 and 0.22 seconds respectively. For the proposed approach, the resulting number of test patches is $468$ yielding $\bfY \in \mathbb{R}^{729\times 468}$, and the dictionary tested is $\bfD \in \mathbb{R}^{729 \times 100}$. The experiments were conducted on ACER core-i3-2.0 GHz processor laptop with 8 GB RAM. Although the proposed approach provides slightly higher computation time, it crucially brings the benefit of providing higher detection performance with respect to the other three methods.
\vspace*{-0.32cm}
\section{Conclusion and Future Work}
\label{sec:Conclusion}
\vspace*{-0.22cm}
In this work, we have demonstrated the performance of an unsupervised approach for bacterial detection in OEM images of distal lung tissue using targeted SmartProbes. We learned a dictionary for background image structure (elastin, collagen, etc.), which was then used to predict any deviating outliers in testing frames. We have provided simulations on two ovine lung datasets instilled with bacteria, which demonstrated that the estimated bacterial counts correlates with the bacterial counts performed by a clinician and good AUC were achieved. However, precautions should be considered when learning the dictionaries for such problems. While annotating ground truth, it is highly likely that the annotator makes mistakes: they can either falsely annotate a bacterium when it is noise, or simply miss-annotating a bacterium due to their overwhelming numbers in each frame. These types of error are common in any annotation process, but it might have a more severe impact on learning the dictionary since our target objects are `dots' with similar structure. Therefore, wrongly annotated/un-annotated bacteria can provide biased dictionary atoms that cause errors in the estimation process. Current investigations include robust methods for learning the dictionary in the case of absence or unreliability of annotations. Moreover, while a standard $\ell_1$- sparsity penalty (in the image domain) is used for the bacterial contributions, structured sparsity and tailored bacteria dictionaries can be used to account for bacteria size/shape and enhance the robustness of the proposed approach. Extension to detection of different bacteria types using colour/multispectral images is also worth investigating.

\bibliographystyle{IEEEtran}
\bibliography{mybib_filtered_noDublicates}

\end{document}